\documentclass[10pt,twocolumn]{article}

\usepackage{cvpr}
\usepackage{times}
\usepackage{epsfig}
\usepackage{tabularx,ragged2e,booktabs}
\usepackage{graphicx}
\usepackage{amsmath}
\usepackage{amssymb}
\usepackage{epstopdf}
\usepackage{csvsimple}
\usepackage{enumitem}

\usepackage[pagebackref=true,breaklinks=true,colorlinks,bookmarks=false]{hyperref}

\cvprfinalcopy % *** Uncomment this line for the final submission

\def\cvprPaperID{3} % *** Enter the CVPR Paper ID here

\ifcvprfinal\fi
\begin{document}

%%%%%%%%% TITLE
\title{Continuous Video to Simple Signals for Swimming Stroke Detection with Convolutional Neural Networks}

\author{
Brandon Victor
\and
Zhen He
\and
Stuart Morgan
\and
Dino Miniutti
\and
Department of Mathematics and Computer Science, La Trobe University, Australia\\
Australian Institute of Sport\\
{\tt\small \{b.victor,z.he,s.morgan\}@latrobe.edu.au}\\
\tt \small Dino.Miniutti@ausport.gov.au
}

\maketitle
%\thispagestyle{empty}

%%%%%%%%% ABSTRACT
\begin{abstract}

In many sports, it is useful to analyse video of an athlete in competition for training purposes. In swimming, stroke rate is a common metric used by coaches; requiring a laborious labelling of each individual stroke. We show that using a Convolutional Neural Network (CNN) we can automatically detect discrete events in continuous video (in this case, swimming strokes). We create a CNN that learns a mapping from a window of frames to a point on a smooth 1D target signal, with peaks denoting the location of a stroke, evaluated as a sliding window. To our knowledge this process of training and utilizing a CNN has not been investigated before; either in sports or fundamental computer vision research. Most research has been focused on action recognition and using it to classify many clips in continuous video for action localisation.

In this paper we demonstrate our process works well on the task of detecting swimming strokes in the wild. However, without modifying the model architecture or training method, the process is also shown to work equally well on detecting tennis strokes, implying that this is a general process.

The outputs of our system are surprisingly smooth signals that predict an arbitrary event at least as accurately as humans (manually evaluated from a sample of negative results). A number of different architectures are evaluated, pertaining to slightly different problem formulations and signal targets.

\end{abstract}

%%%%%%%%% BODY TEXT
\section{Introduction}

\begin{figure}
   \centering
   \includegraphics[width=\columnwidth]{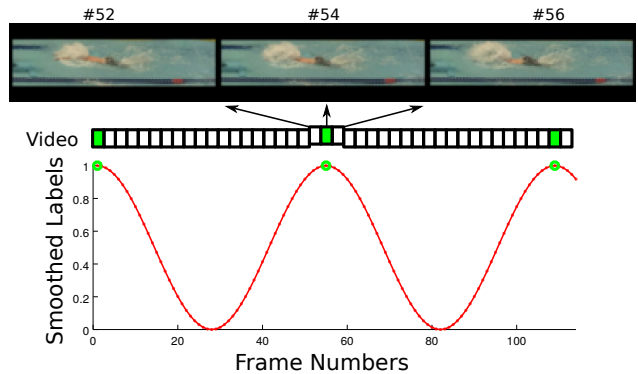}
   \caption[Label Smoothing]{The raw labels for swimming stroke detection are very sparse. To train a CNN on these labels, we smooth them to produce a continuous target signal instead of discrete binary values.}
   \label{fig:smoothing}
\end{figure}

Automatic video analysis of sports boasts several attractive features: it can be done quickly, with a simple camera, objectively, without obstructing the athletes in any way and without using the time of sports analytics experts. Stroke rate is an important metric used in swimming training, and currently, experts spend a significant amount of time manually labelling each stroke in a video in order to provide statistical feedback to the swimmers. We call this task \textbf{discrete event detection} (distinct from, \textbf{event detection}; which is detecting the beginning and end of an action).

There has been much research in extracting useful information from video and recent improvements in training deep CNNs \cite{batchNormalisation,ResNet} allow them to replace whole sections of computer vision pipelines for video analysis \cite{Zecha2015,Zecha2017,Wang2016,3DCNN}. In this paper, we describe a method to train a simple CNN for \textbf{discrete event detection}.

\newcolumntype{L}{>{\RaggedRight\arraybackslash}X}
\begin{table*}
   \caption[Video Analysis tasks]{Summary of related, but distinct, video analysis tasks discussed in this paper.}
   \begin{center}
      \begin{tabular}{| p{0.16\linewidth} | p{0.35\linewidth} | p{0.24\linewidth} | p{0.16\linewidth} |}
         \hline
         \bfseries {Name} & \bfseries {Description} & \bfseries {Example Output} & \bfseries {Public Datasets}
         \tabularnewline\hline
         Action Recognition & Classifying a whole non-continuous video as a particular action & This video shows `soccer' & UCF\cite{Soomro2012}, Sports-1M\cite{fusionMethods} \tabularnewline\hline
         Action Localisation / Event Detection & Locating any number of actions in continuous video and classifying them & There was a `horse riding' action from frames $a$ to $b$, etc. & THUMOS\cite{jiang2014thumos}, TRECVID MED\cite{2016trecvidawad} \tabularnewline\hline
         Discrete Event Detection & Determine precise frame numbers that an event occurs & The swimmer's hand entered the water on frames $\left\{a,b,...\right\}$ & To our knowledge: none \tabularnewline\hline
      \end{tabular}
   \end{center}
   \label{tbl:summaries}
\vspace*{-\baselineskip}
\end{table*}

% . \textbf{Action recognition}: classifying a non-continuous video or clip, \eg given a clip of soccer, we want the algorithm to tell us that it is soccer. In the figure, we show that there is a single classification label for a whole video. UCF101\cite{Soomro2012} and Sports-1M\cite{fusionMethods} are public datasets for action recognition. \textbf{Action localisation}: localising and classifying actions in untrimmed video, \eg given a video including someone playing soccer and then going base jumping, we want the algorithm to tell us that there is someone playing soccer from frame $y_1$ to frame $y_2$, and also that there is someone base jumping from frame $y_3$ to frame $y_4$. In the figure we show that this is a prediction for the start and end for each action, along with a classification. THUMOS\cite{jiang2014thumos} and TRECVID MED\cite{2016trecvidawad} are public datasets used for action localisation. \textbf{Action localisation} is often referred to as \textbf{event detection}; we choose the first name to clearly differentiate between this and \textbf{discrete event detection}. \textbf{Discrete event detection}: determining precise frame numbers of events, \eg which frames did the swimmer's hand enter the water.

CNNs are constrained to fixed-size input/output, requiring a sliding window approach. The naive approach would be to classify each window as a \textbf{stroke} (denoted as 1) or \textbf{not a stroke} (denoted as 0), but training a CNN on these labels directly is an \textit{unstable learning problem}. First, there is a huge imbalance between positive and negative examples; statistically speaking, always predicting 0 is very accurate, and hence favoured by a machine learning algorithm. Second, as Figure \ref{fig:smoothing} shows, the neighboring frames of each `stroke' have similar pixel contents, but would have different labels; there is very little correlation between pixel content and the desired output. This second problem is exacerbated by the ambiguity inherent to frame-specific labels (it is sometimes unclear even to human experts).

Our key contribution is to translate this \textit{unstable learning problem} into a more numerically optimal learning problem. First we translate the raw labels to a continuous signal. Figure \ref{fig:smoothing} shows that by forcing nearby frames to have similar targets, we also create many non-zero labels, solving both problems at once, and introducing a tolerance to inconsistent label positions. Next, we train a CNN to match this continuous signal. Finally, from the predicted continuous signal we discretise the signal back into precise frame numbers.

\begin{figure}
   \centering
   \includegraphics[width=\columnwidth]{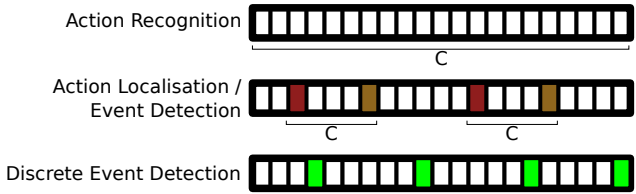}
   \caption[Terminology]{Comparing different video tasks: the cells represent video frames, colored cells means a frame number annotation, and `C' means a classification annotation for some number of frames.}
   \label{fig:terminology}
\end{figure}

For clarity, we disambiguate between three important, and related, video analysis tasks: action recognition, action localisation/event detection and discrete event detection. Table \ref{tbl:summaries} describes each task with examples/public datasets. Figure \ref{fig:terminology} shows the targets for each task with respect to video frames. To put \textbf{discrete event detection} (the focus of this paper) more formally: we want some function $g$ that processes some video $\mathbf{V}$ with $N$ frames into frame numbers $\mathbf{F}$ (denoting where events occurred) such that

\begin{equation}
g(\mathbf{V}) = \mathbf{F} \;,\;\; with \;\;\; \left(\forall i\in\mathbf{F}\right)\;i \in [1,N].
\label{eq:eventDetection}
\end{equation}

We have conducted several experiments to test different ways of using CNNs for discrete event detection. The key findings are:
\begin{itemize}[noitemsep]
\item using a CNN to predict strokes works extremely well in both swimming and tennis (F-Score = 0.92 and 0.97, respectively, at 3-frame tolerance) with no domain-specific settings, suggesting this training process is general to many sports.
\item softening the targets to produce a smooth 1D signal is more effective than using hard labels.
\item differentiating between swimming styles is not important for swimming stroke detection.
\item early fusion of video frames with a 2D architecture is able to discern motion information and out-performs a single frame.
\item there is very little variance between the model's raw and smoothed signal output ($\sim$3\% on average).
\end{itemize}

\section{Related Works}
There are two backgrounds to consider for this work. The first is how generic image classification, video action recognition and action localisation have evolved recently with deep learning. The second is swimming analysis and stroke detection. To our knowledge, our specific definition of \textbf{discrete event detection} has not been studied in generic video analysis research.

\subsection{Generic image classification and video analysis tasks}

Video action recognition could be described as image classification using temporally unbounded image frames as input. Solutions to image classification are often adapted and extended to work for video data \cite{twoStream}. The typical pipeline for video action recognition with hand-crafted features uses a Bag of Visual Words approach and has up to 5 stages; feature extraction, feature pre-processing, codebook generation, feature encoding, and pooling, before going through a classification algorithm \cite{Peng2015}. Research on hand-crafted features focuses on improving part of this pipeline (usually feature extraction or feature encoding).

The most well-known algorithms for feature extraction are: Scale Invariant Feature Transform (SIFT) \cite{Lowe2004}, Histogram of Gradients (HOG) \cite{Dalal2005}, Histogram of Optical Flow (HOF) \cite{Chaudhry2009}, and Motion Boundary Histograms (MBH) \cite{Wang2013a}. For feature encoding, the best performing methods are Fisher Vectors (FV) \cite{Perronnin2010} or variations, like Vector of Locally Aggregated Descriptors (VLAD) \cite{Jegou2012}.

Until recently the state-of-the-art in video action recognition has been hand-crafted feature extraction. Oneata \etal \cite{Oneata2013} achieved state-of-the-art performance on the TRECVID MED dataset using densely extracted SIFT and MBH for feature extraction, FV for feature encoding and an SVM for classification. Gaidon \cite{Gaidon2013} suggested that each action consists of multiple parts, called `actoms'. By modelling actions as a series of actoms it simplifies the recognition of parts, but requires denser labels, which is usually not available. Peng \etal \cite{Peng2015} achieved state-of-the-art mAP on the UCF101 dataset with an almost exhaustive search across combinations of known algorithms for each step of the above pipeline.

Image classification competitions are currently dominated by CNNs \cite{AlexNet,VGG,ResNet}. These have provided very large improvements over methods involving hand-crafted visual features; from 25.8\% error in ILSVRC2011 using compressed Fisher Vectors to 3.57\% error in ILSVRC2015 \cite{ResNet,Russakovsky2015}.

Like the hand-crafted features before them, Convolutional Neural Networks have since been adopted for use in video action recognition. Neural networks tend to function on fixed-size input and output, so it is not immediately obvious how to use CNNs for video. Wang \etal \cite{Wang2015} evaluated images frames with a pre-trained VGG model \cite{VGG} to produce frame-level features from the activations of later layers. They averaged and encoded these with VLAD before using cosine similarity for classification. Simonyan and Zisserman used their VGG\cite{VGG} architecture in a two-stream architecture \cite{twoStream} using image frames and optical flow to improve on state-of-the-art. They make action predictions at 25 equidistant frames in each clip and evaluate for a whole clip by using the last layers' activations in an SVM.

Karpathy \etal \cite{fusionMethods} evaluated different methods of processing small fixed-size windows with CNNs for action recognition: single frame, early fusion, late fusion and 3D convolutions. They evaluated by simply sampling several locations in videos and pooling the results by majority voting. Tran \etal \cite{3DCNN} used a 3D Convolutional Neural network to produce 16-frame clip-level descriptors in a sliding window approach, averaging across the whole video to produce a video-level descriptor which was classified with an SVM.

To our knowledge, the best result on UCF101 is currently by Wang \etal \cite{Wang2016}. They learn a transformation matrix per action class, from two descriptors obtained at statistically likely locations for the beginning and end of an action with siamese networks, selecting the transformation that minimised the distance.

% They use siamese networks evaluated at statistically likely frames to produce two descriptors, interpreted as describing the beginning of the action and the beginning of the effect, respectively. They learn a transformation matrix from the output of one to the other, per action class, and at evaluation time, they select the action transformation that minimises the distance between the second descriptor and the transformed descriptor.

\subsubsection{Action Localisation}

\begin{figure}
   \centering
   \includegraphics[width=\columnwidth]{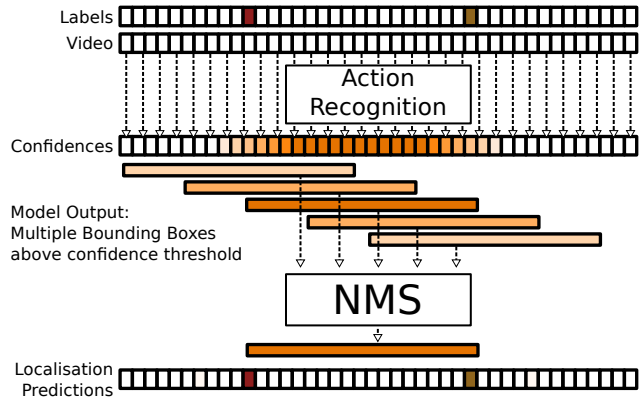}
   \caption[NMS in action]{Most other work treats action localisation as a classification problem. A near-perfect action recognition algorithm would produce several bounding boxes around each ground truth; the confidence would smoothly increase as the bounding box overlaps more with the original labels. Typically this granular information is not leveraged and NMS is used to select a single bounding box.}
   \label{fig:nmsInAction}
\end{figure}

The main reason we differentiate between action localisation and discrete event detection is to make clear that discrete event detection requires much more precise predictions. Action localisation tends to be thought of as action recognition on untrimmed video, where subsets of frames need to be classified. This introduces a problem if the subset of frames are interpreted as a bounding box through time; there are too many positive predictions. From this interpretation, it seems natural that the most applicable bounding box will have the highest confidence and nearby predictions should be ignored. The most common solution is non-maximal suppression \cite{Gaidon2013,Oneata2013,actionLocalisation}. Gaidon \etal \cite{Gaidon2013} and Oneata \etal \cite{Oneata2013} used their action recognition for action localisation with a sliding window, using non-maximal suppression to obtain distinct predictions. Shou \etal \cite{actionLocalisation} explicitly separated detecting an action and classifying it, creating separate 3D CNNs to do each job. They produced classifications for a small clip at a time, in a sliding window over the videos, with non-maximal suppression to produce the final predictions.

Yet, it can be seen that the expected output of a near-perfect action recognition algorithm would approximate the smoothed target labels proposed in this paper. Consider that the confidence in a bounding box using such an algorithm will vary smoothly from 0\% recognition for no overlap to 100\% recognition for complete overlap around each true action (this is shown in Figure \ref{fig:nmsInAction}). In a sense, the predictions for both \textbf{action localisation} and \textbf{discrete event detection} are a continuous signal which need to be discretised. While others have noted that positive classifications that are temporally close to one another can be utilised to make predictions more accurate \cite{Ke2005,Zecha2017}, this is only mentioned as a side-note. To our knowledge no one else directly learns and utilises this natural continuous signal.

\subsection{Swimming Analysis}

Sha \etal \cite{Sha2013} used the differences in colour between the water and the foreground objects (\ie swimmers and lane ropes) to spatially locate swimmers in untrimmed video. They first isolate the lanes, and then the swimmers themselves, building an image mask for the swimmer. They were quite successful at tracking swimmers (88.6\% of frames after post-processing), but the method does not translate to other sports at all. They did not attempt to detect strokes in this work.
% They manually described the possible states of a swimming race (start, dive, under water, turn, normal swimming, end) and applied different logic in each stage.

Building on their work in spatially localising the swimmer, Sha \etal \cite{Sha2014} predicted swimming stroke rate by locating the swimmer's elbow with a deformable parts model, tracking it's y-position through time to produce a noisy, periodic pattern. They used a relatively small set of videos (freestyle only) and noted difficulties with varying illumination, camera angle and zoom. They obtained an average of 5\% error in stroke rate. By this metric, our method achieves ~0\% error (see the `10+' column in Figure \ref{fig:cumulativeFrameDist}).

Tong \etal \cite{Tong2006} used broadcast video of swimming races to predict the style. They used a lengthy pipeline of processing involving a few custom hand-crafted features used in a neural network and SVM, evaluated at multiple frames, with majority voting to aggregate.

Zecha \etal \cite{Zecha2017} detected swimming strokes using video taken through a glass wall, showing the whole swimmer above and below the water. They used an AlexNet architecture \cite{AlexNet} to classify patches of an image to locate joints at each frame, from which they used a deformable parts model to create a pose estimation. They fed the pose estimations into a neural network to predict several types of swimming event. Compared to our process: their videos are constrained to a lab setting where the swimmer's whole body is visible at all times, while ours uses natural video taken at races; and theirs is not an end-to-end deep learning solution.

\section{Modified problem description}
\begin{figure}
   \centering
   \includegraphics[width=\columnwidth]{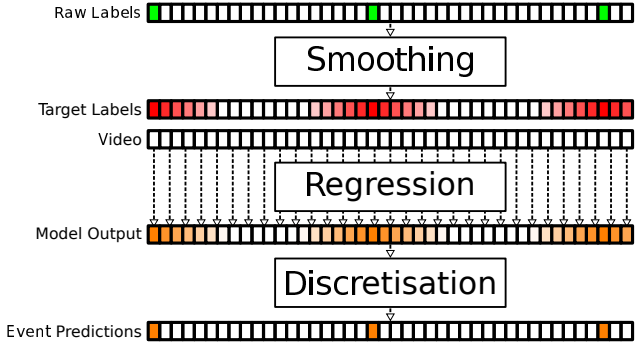}
   \caption[Two parts to the problem]{We define two parts to our solution of the modified problem for discrete event detection. We use a CNN to do the complex mapping from video frames to a simple target signal (regression), and simple thresholding to predict precise frame numbers (discretisation).}
   \label{fig:twoParts}
\end{figure}
Equation \ref{eq:eventDetection} defines the task of discrete event detection. As explained in the introduction, directly classifying each window is an unstable learning problem. Hence we break the problem into two parts: regression and discretisation (see Figure \ref{fig:twoParts}). The regression part is a mapping from a small number of video frames to a point on a signal. The target signal is the result of label smoothing (see Figure \ref{fig:labelTypes} and Section \ref{sec:labels}).

\textbf{Regression.} Let $y_i$ be the smoothed label for the i-th frame with $\mathbf{x}_i$ the input video frame. Then, we want some function $R$ that produces an estimate $\hat{y}_i$ such that

\begin{equation}
	R(\{\mathbf{x}_{i-w}, ..., \mathbf{x}_i, ..., \mathbf{x}_{i+w}\}) = \hat{y}_i
\end{equation}
with the minimum mean-squared error loss
\begin{equation}
	\min \sum_{i=w}^{(N-w)} \left(  y_i - \hat{y}_i  \right)^2\;.
\end{equation}

Where $(2w+1)$ is the width of the window of frames used as input and N is the number of frames in the video.

\textbf{Discretisation.} From the estimated signal $\hat{\mathbf{y}}$, we want some function $D$ that produces specific frame numbers $\hat{\mathbf{F}}$;

\begin{equation}
	D(\hat{\mathbf{y}}) = \hat{\mathbf{F}} = \{\hat{f}_1, \hat{f}_2, ..., \hat{f}_m\}.
\end{equation}

Where $\hat{f}_i$ is a stroke prediction (frame number) and $m$ is the number of predicted strokes.

\section{Our solution}

\begin{figure}
   \centering
   \includegraphics[width=\columnwidth]{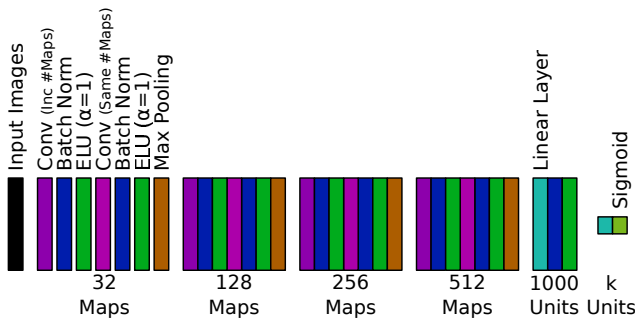}
   \caption[Architecture]{The base architecture used for all experiments, using the ELU non-linearity\cite{ELU} and Batch Normalization\cite{batchNormalisation}.}
   \label{fig:architecture}
\end{figure}
\textbf{Regression.} We used a standard CNN for the regression function $R$. As a base architecture, we use a CNN loosely based on VGG-B\cite{VGG}; a pattern of blocks of two convolutions with a max pooling layer on the end, however the number of maps, blocks and fully connected layers are different (see Figure \ref{fig:architecture}). The kernel sizes for all convolution layers are 3x3 (2D case) or 3x3x3 (3D case) with padding to retain image size. The max pooling layers use a 5x5 spatial kernel with stride of 2x2 to downsample and padding similar to convolution layers. For the 3D case, the max pool layers have a stride of 1 in the temporal dimension and no padding if the intermediate temporal dimension is larger than 3.

\textbf{Discretisation.} Our discretisation process involves three steps. First, we smooth the signal with a weighted moving average, producing a smoother curve. Second, we threshold the signal at the mean (for tennis we threshold at 0.5), producing a square wave. Third, we scan linearly through the signal and for every unbroken chain of 1's, we declare the middle frame of that chain to be a predicted stroke $\hat{f}_i$.

\section{Exploratory Factors}

We consider several different CNN architectures. First we consider how temporal information should be incorporated into the model. Second we consider different ways to use swimming style information, necessitating architecture changes and having implications for the model capacity and generalisation of the result.

\subsection{Using Temporal Data}
\label{sec:temporalData}
We compared `single frame', `early fusion' and 3D CNN as in \cite{fusionMethods}. Taking a single frame as input is used as a baseline technique. In early fusion we stack the frames together along the maps dimension and use a 2D CNN. A 3D CNN has three dimensional convolution kernels; where a 3D CNN retains temporal representations throughout, a 2D CNN does not and hence is less equipped to find motion correlations. Temporal/motion information is especially important for the case of occlusions. A single frame model is unable to cope with a mostly occluded frame, while the others will be able to use the surrounding information to make some kind of prediction. A 3D CNN has more parameters; increasing model capacity and computation time required.

\subsection{Using Swimming Style Data}
\label{sec:swimmingStyleData}
We wanted to determine the most effective way to use classes for discrete event detection. In this paper, we treat the different styles as a proxy for different classes of action. Four styles were used: Backstroke, Breaststroke, Butterfly and Freestyle.

A neural network model is an approximation of an unknown target function. The potential benefit of adding complexity to the target function is that it generalises better with a small cost to model capacity, resulting in better overall performance. The potential detriment is that it either does not generalise any better and/or has a large cost of model capacity which will hurt overall performance.
% A model's capacity is the complexity of function that it can approximate and/or how well it can approximate that function. So, requiring that a model performs more tasks increases the complexity of the target function, reducing how well it can approximate the function. It is designed to pick up on correlations in data, even correlations that do not generalise, so choosing a more difficult unknown target function may cause the model to require a more generalised understanding of the task and improve generalisation performance.

\begin{figure}
   \centering
   \includegraphics[width=\columnwidth]{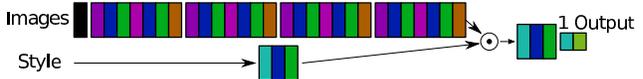}
   \caption[Style as Input Architecture]{A method to use the style as extra input for the network. The input image frames go through the convolution section like normal, while the style (one-hot encoded) is processed in parallel to produce a gating vector on the output of the convolution section.}
   \label{fig:styleAsInputArchitecture}
\end{figure}
\textbf{Model per style.} The simplest target function investigated here was to limit our dataset to just one style of video (freestyle; $\sim$50\% of the videos) for both training and evaluation with just one output number ($k=1$). Even if this method works slightly better, it is less practical to construct a separate model for each class of action, and is included more as a comparison point than a viable alternative.

\textbf{All styles.} Conversely, we use $k=1$, and feed \textit{all} videos to the model. Given the subtleties between the styles, it is plausible that this change introduces a small increase in complexity with a relatively larger generalisation effect.

\textbf{Multi-class.} By using $k=4$, one for each style, we can predict and learn for each style separately. This is based on the idea of multi-task learning. Let $u_i$ be the scalar target for the i-th frame used for the case of $k=1$, and let $\mathbf{s}$ be the style (class), one-hot encoded. Then the targets for the `multi-class' model are
\begin{equation}
   \mathbf{y}_i=u_i\mathbf{s} .
\end{equation}
Note that since $\mathbf{s}$ has three zeros, 3 of the values in $\mathbf{y}$ are always zero.

\textbf{Style as Input.} We also experiment with providing the class label (one-hot encoded) as extra input with $k=1$, called a `style as input' model. After passing the class label through a single linear layer to match sizes, the class label is element-wise multiplied with the flattened output of the convolutional section of the network (see Figure \ref{fig:styleAsInputArchitecture}). Thus the network learns a gating on the convolution features, based on the style. This reduces the complexity as compared to `all styles', while retaining the concept of explicitly separating the swimming styles from the `multi-class' model.

\subsection{Target Signal}
\label{sec:labels}
\begin{figure}
   \centering
   \includegraphics[width=\columnwidth]{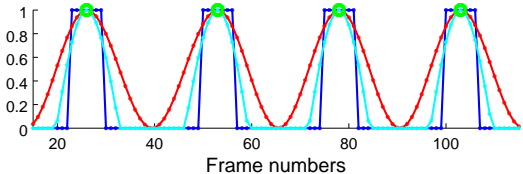}
   \caption[Label Transformations]{A visual comparison between the different ways of transforming the original labels (shown as green circles) into smoother functions used in this paper (red, cyan and blue lines; `sine', `truncated sine' and `square' respectively).}
   \label{fig:labelTypes}
\end{figure}

We propose to compare three different smoothing functions to create the target signal; called `square', `sine' and `truncated sine' (see Figure \ref{fig:labelTypes}). The labels for frames are - conceptually - modified by their proximity to events. Any transformation needs to account for events potentially being very close together and being very far away from each other; e.g. a fixed shape must not be wider than the smallest distance between events. With regards to swimming, the number of frames between the strokes is indicative of how quickly the swimmer performs the action, which is not consistent across time, videos or style. Additionally there are periods of no strokes where the swimmer turns around at the edge of the pool.

For the `square' labels, if the frame is within 3 frames (inclusive) of an event frame, then it is labelled as a 1. For the `sine' and `truncated sine' labels the first step is to fit a cosine between every pair of strokes (except through a swimmer's turn, where the average cosine is fit to the edges, leaving most of the turn action labelled as 0). Let $\mathbf{c}$ be these intermediate labels from the first step, with $c_n$ being the intermediate label for the $n$-th frame. Then the transformed labels are:
\begin{equation}
   y_n = \max\left\{ac_n+(1-a), 0\right\}.
\end{equation}
Where `sine' labels use $a=\frac{1}{2}$ and `truncated sine' labels use $a=1$.

For the tennis evaluation, the strokes were more sparse. A fixed shape was constructed similarly around each label, with $\mathbf{c}$ representing half of a cosine wave with a period of 40 frames, the peak centred on the stroke.

\section{Implementation details}
The model parameters were initialised with He \etal's method in \cite{ResNet}. The Adadelta\cite{adadelta} optimiser was used - thus no learning rate was selected - with a minibatch size of 64. As this is now a regression problem, the loss function used is Mean-Squared Error. No regularisation of weights was used. All frames' pixels were encoded in the YUV colour-space and downsampled to 128x48. The video frame pixel values were standardised at the channel level with the mean and standard deviation across the whole dataset. The resulting data was augmented by random zooming (up to 20\% larger, cropping back to original size), and random colour-space variation (between 1/3x and 3x scaling, applied per-channel).

By skipping input frames, a lower fps can be simulated, introducing more motion between frames. Along with temporal window size, this introduces some hyper-parameters for using video. There were some small experiments done to determine the best values for these; the number of frames appeared mostly irrelevant, but skipping frames was found to be strictly worse beyond a certain point. Every second frame was skipped for the swimming experiments in Section \ref{sec:Exp}, but not for the tennis experiment as they were only 30fps video (as opposed to the swimming videos at 50fps).

To minimise the disk I/O, the input frames are pulled in succession from the videos, from beginning to end, caching the previous frames for each video. The video from which the frames are pulled is chosen at random from a discrete non-uniform distribution of the number of frames in each video (\ie a video with 2\% of the total number of frames has a 2\% chance of being selected). With the number of videos present, there is typically no bias in a minibatch.

\subsection{Datasets and Video preprocessing}
\label{sec:datasets}
We used two datasets which were hand-labelled by experts at the Australian Institute of Sport. The swimming dataset was the initial focus, while the tennis data was the most readily available different dataset that could be used as a validation that the training process was general across sports.

\textbf{Swimming data.} The dataset used for the experiments - unless otherwise stated - contains ~15k labelled swimming strokes in ~650k frames of video (at 50fps) at two venues, consisting of 40 different swimmers. Unprocessed video of swimming races can include multiple swimmers, but by indicating only one set of strokes, and there was no guarantee to which swimmer the labels belong. Thus these videos are the result of being preprocessed as in \cite{Sha2013} to extract the lanes from colour information: cropped and sheared to obtain a single swimmer's lane that is axis-aligned. We used $\sim$80\% of the data for training. A `stroke' is defined as ``the frame that the swimmer's hand enters the water''

\textbf{Tennis data.} Tennis is a good sport to test the generality of our method since the background, colours and human motions are very different from swimming while the moment the key event occurs is similarly clear in tennis (racket hits ball). This dataset consisted of ~1.3k labelled tennis strokes in ~270k frames of video (at 30fps) at two venues with of 4 different players practising tennis shots. Swimming strokes appear much more densely than the tennis strokes due to their periodic nature and there was a relatively large number of tennis strokes that were unlabelled. Thus the dataset of videos for tennis was constructed as small clips around each labelled stroke, and large sections of video with no strokes used as background videos. The sliding window evaluation used in this paper means that the network uses the same frames and thus is still completely applicable to the original videos. We used $\sim$80\% of the tennis data for training, and $\sim$20\% for testing.

\section{Experiments}
\label{sec:Exp}
% As a sanity check, a model was trained with non-transformed labels (many 0's, few 1's); the model learned to always output 0 almost immediately and did not learn anything useful for generalisation.

For the swimming experiments the default decisions are to use early fusion of input, `style as input', and to train using `sine' transformed labels. The input for all early fusion and 3D architecture models for swimming is 11 frames wide (5 frames wide for tennis). The decisions are compared with one another on the validation set, as there is no publicly available dataset to compare performance on.

\subsection{Evaluation Metrics}
The main evaluation metric used is the $F_1$-score:
\begin{equation}
   F_1\text{-score}=2\times\frac{\text{Precision}\times\text{Recall}}{\text{Precision}+\text{Recall}},
\end{equation}
henceforth called the F-score. Each stroke prediction was considered a true positive if it was within 3 frames of any initial stroke label. All other stroke predictions were considered false positives, and all initial stroke labels that were not covered were considered false negatives. The stroke predictions were not evaluated frame-to-frame because the precise frame number is ambiguous, even to human experts, and the initial labels have approximately this same margin of error as well.

We use two additional metrics for further comparison. \textbf{Average frame distance}, measuring the average distance between each predicted stroke $\hat{f}_i$ and the nearest true stroke $f_j$ (denoted `$\min\left|\hat{f}_i-f_j\right|$'). And \textbf{average difference to smoothed}, measuring the average point-wise difference between the raw output signal $\hat{\mathbf{y}}$ and the smoothed output signal (denoted `$\Delta$ Smooth').

\begin{table*}
   \caption[Results]{Experimental results by Temporal Architecture, Style Input Mode, and Target Signal. The default model of `Early Fusion'/`Style As Input'/`Sine' is mentioned once per group for the sake of easier comparison. The models in bold obtained the best F-Score in each section.}
   \begin{center}
      \begin{tabular}{| c | c | c | r | r | r |}
         \hline
         \bfseries {Temporal Architecture} & \bfseries {Use of Style Data} & \bfseries {Target Signal} & \bfseries {F-Score} & \bfseries {$\min\left|\hat{f}_i-f_j\right|$} & \bfseries {$\Delta$ Smooth}
         \begin{filecontents*}{csv/results.csv}
theflag, Use of Temporal Data,Use of Style,Label Transformation,F-Score,Frame Dist, delta Smoothed
X,None (Single Frame),Style As Input,Sine,0.894,1.690,0.0391
,Early Fusion,,,0.900,1.711,0.0340
,\textbf{3D Conv},,,\textbf{0.922},1.604,0.0330
X,Early Fusion,\textbf{All Styles},Sine,\textbf{0.906},1.722,0.0356
,,Model per Style,,0.875,1.696,0.0419
,,Multi-Class,,0.863,1.811,0.0331
,,Style as Input,,0.900,1.711,0.0340
X,Early Fusion,Style As Input,Sine,0.900,1.711,0.0340
,,,\textbf{Truncated Sine},\textbf{0.912},1.587,0.0419
,,,Square,0.879,1.960,0.0660
\end{filecontents*}
\csvreader[
               head to column names,
               before line=\ifthenelse{\equal{\theflag}{}}{\tabularnewline}{\tabularnewline\hline},
            ]{csv/results.csv}{}{
               \csvcolii&\csvcoliii&\csvcoliv&\csvcolv&\csvcolvi&\csvcolvii
            }
         \tabularnewline\hline
      \end{tabular}
   \end{center}
   \label{tbl:results}
\vspace*{-\baselineskip}
\end{table*}

\begin{figure}
   \centering
   \includegraphics[width=\columnwidth]{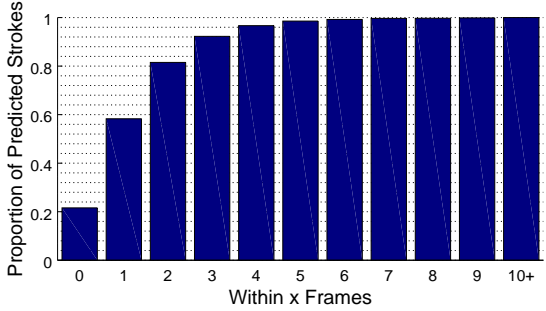}
   \caption[Cumulative Histogram of Frame Distances]{The 3D CNN from the swimming experiments (the model with the best results) completely missed very few strokes. This shape is representative of all models.}
   \label{fig:cumulativeFrameDist}
\end{figure}
\begin{figure}
   \centering
   \includegraphics[width=\columnwidth]{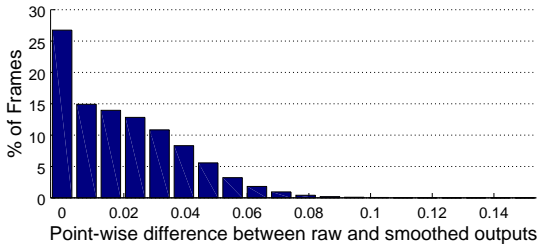}
   \caption[Histogram of Deltas with Smooth]{A typical histogram of the point-wise difference between the raw outputs and the smoothed output (this is from the 3D CNN).}
   \label{fig:deltaSmoothHist}
\end{figure}

\subsection{Using Temporal Data}
\begin{figure}
   \centering
   \includegraphics[width=\columnwidth]{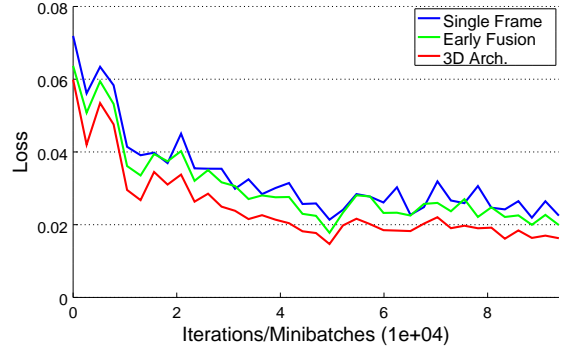}
   \caption[Temporal Data Loss]{The validation loss between the different methods for including temporal data (minibatch size of 64).}
   \label{fig:temporalDataLoss}
\end{figure}
In Section \ref{sec:temporalData}, we mention three ways of including temporal data. Although the 3D convolution architecture obtained a higher F-score (Table \ref{tbl:results} and Figure \ref{fig:temporalDataLoss}), it takes several times longer and more parameters than the others.

Using a single frame does quite well, however, sometimes the swimmer will be occluded by foreground objects, or failed pre-processing. There are not many occlusions in the dataset, and using early fusion does not introduce many new parameters, so the increase in performance from no temporal information to early fusion can be attributed to finding a better representation using motion information. It is unclear what proportion of the improvement from early fusion to a 3D architecture can be attributed to an intrinsic understanding of motion by convolving through time, and how much is simply because a 3D architecture has more parameters.

\subsection{Using Swimming Style Data}
As mentioned in Section \ref{sec:swimmingStyleData}, there are four ways to use the style of swim as classes. Of these, the simple `all styles' was the best (Table \ref{tbl:results}). The benefit of more videos must outweigh the complexity added by including all styles in a single model. There must be enough information across styles that there is no benefit in being provided the style.

It is interesting that the `multi-class' model obtained a markedly lower F-score. The targets for each input was a 4-element vector, consisting of three 0s and a transformed target value. We speculate that this method of training may have been a less stable training objective because the network must identify features that are both distinctive to the style \textit{and} the proximity to the original stroke at the same time. This actually creates less incentive to share features across styles, since it must find features specifically to differentiate completely between the styles. At evaluation time, the style is inferred by finding the output index with the highest sum of outputs over a video. The models did not incorrectly identify any styles.

\subsection{Producing Different Signals}
A surprising result is how smooth the outputs are in all cases; Table \ref{tbl:results} shows that the average difference between the raw output signal and the smoothed version is extremely small (see Figure \ref{fig:deltaSmoothHist}).

\subsection{Method applied to Tennis}
\begin{table}
   \caption[Tennis]{When the training method is applied to Tennis, we obtain at least as good results as for swimming.}
   \begin{center}
      \begin{tabular}{| c | r |}
         \hline
         \bfseries {Temporal Architecture} & \bfseries {F-Score}
         \begin{filecontents*}{csv/tennisResults.csv}
Use of Temporal Data, F-score
Single Frame, 0.964
Early Fusion, 0.977
\end{filecontents*}
\csvreader[
               head to column names
            ]{csv/tennisResults.csv}{}{
               \tabularnewline\hline\csvcoli&\csvcolii
            }
         \tabularnewline\hline
      \end{tabular}
   \end{center}
   \label{tbl:tennisResults}
\vspace*{-\baselineskip}
\end{table}

As described in Section \ref{sec:datasets}, the input videos were preprocessed to exclude the unlabelled tennis strokes. The input data frames were downsampled to 192x128 (larger than the swimming videos) because more of the image was taken up by background. All other training settings and data augmentation was identical. A `single frame' and a `early fusion' model were trained from scratch on this dataset (both models were also `all styles' and used `truncated sine' labels). Both achieved better results as compared to equivalent models for swimming (see Table \ref{tbl:tennisResults}), however this may be due to more precise labels (see Section \ref{sec:manualInspection}).

While the early fusion model did not produce any false positives further than 6 frames away from a expected stroke, the single frame model had several false positives where a player was simply walking across the court. Without the motion information, the model was unable to tell the difference between a tennis racket being held in front of the player and the tennis racket being used in a swing.

\subsection{Manual Inspection of False Predictions}
\label{sec:manualInspection}
For the best performing model from the swimming experiments, by randomly inspecting false predictions we noted that the majority of the false predictions were more accurate than the hand-annotated labels, so the true F-Score for this task is likely to be much higher than reported. From Figure \ref{fig:cumulativeFrameDist}, it can be seen that among the false predictions, very few were completely missed (or added). We suggest that the majority of false predictions were due to temporally imprecise labelling.

For the best performing model from the tennis experiments, due to the relatively small number of mistakes, we are able to directly classify each mistake. All incorrect predictions were due to temporal imprecision in the model, with only two strokes missed completely. These two missed strokes were dive shots without any back-swing, which are very rare in the dataset. More video of this kind of stroke should allow the model to correctly identify these, as well.

\section{Conclusion}

We have shown that a Convolutional Neural Network can learn to process continuous video into a 1D signal with peaks corresponding to arbitrary events. This holds for both swimming and tennis, implying that it can be used to train separate models for many other sports. Additionally, our results imply that some amount of label smoothing provides a more numerically accurate target.

\section{Future work}

From our evaluations, the performance of swimming stroke detection does not have much room for improvement. Instead, future work would most likely focus around extracting more detailed information from video and incorporating this task into the training process. For example, taking raw video containing multiple swimmers, detecting the location of those swimmers, and producing the stroke estimates for each swimmer.

We would also like to see the effect of this kind of problem translation for action localisation. This process is very naturally extended to detecting bounding boxes for generic action localisation by taking the thresholded signal and treating each unbroken chain of 1s as a bounding box in time rather than discretising to a single point.

{\small
\bibliographystyle{ieee}
\bibliography{bibli}
}

\end{document}